\documentclass{article}
\usepackage{arxiv}

\usepackage{amsmath}

\usepackage{amssymb}
\usepackage{latexsym}
\usepackage{hyperref}
\usepackage{url}
\usepackage{xcolor}
\usepackage{colortbl}
\usepackage{color}

\usepackage{hyperref}
\usepackage{mathtools}
\usepackage{multirow}

\author{
	Shozo Saeki \\
	Center for Information Technology \\
	Ehime University \\
	Matsuyama, Ehime and 790-8577, Japan \\
	\texttt{saeki.shozo.cg@ehime-u.ac.jp} \\
	\And
	Minoru Kawahara \\
	Center for Information Technology \\
	Ehime University \\
	Matsuyama, Ehime and 790-8577, Japan \\
	\texttt{kawahara@ehime-u.ac.jp} \\
	\And
	Hirohisa Aman \\
	Center for Information Technology \\
	Ehime University \\
	Matsuyama, Ehime and 790-8577, Japan \\
	\texttt{aman@ehime-u.ac.jp} \\
}
\title{Combined Hyperbolic and Euclidean Soft Triple Loss Beyond the Single Space Deep Metric Learning}

\newcommand{\xe}[1]{\mathbf{x}_{#1}^{E}}
\newcommand{\xh}[1]{\mathbf{x}_{#1}^{H}}
\newcommand{\pe}[1]{p_{#1}^{E}}
\newcommand{\ph}[1]{p_{#1}^{H}}

\newcommand{\sh}[1]{\mathbf{s}_{#1}^{H}}
\newcommand{\mpe}[2]{\mathbf{p}_{#1, #2}^{E}}
\newcommand{\mph}[2]{\mathbf{p}_{#1. #2}^{H}}
\newcommand{\sime}[2]{\mathcal{S}_{E} ( #1, #2 )}
\newcommand{\simh}[2]{\mathcal{S}_{H} ( #1, #2 )}

\newcommand{\dise}[2]{D_{E} ( #1, #2 )}
\newcommand{\dish}[2]{D_{H} ( #1, #2 )}

\newcommand{\posf}[3]{l^{+}_{#1} \left( #2, #3 \right)}
\newcommand{\negf}[3]{l^{-}_{#1} \left( #2, #3 \right)}
\newcommand{\arctanh}{\mathrm{arctanh}}
\newcommand{\hypc}{\mathfrak{c}}

\begin{document}
\maketitle

\begin{abstract}
Deep metric learning (DML) aims to learn a neural network mapping data to an embedding space, which can represent semantic similarity between data points.
Hyperbolic space is attractive for DML since it can represent richer structures, such as tree structures.
DML in hyperbolic space is based on pair-based loss or unsupervised regularization loss.
On the other hand, supervised proxy-based losses in hyperbolic space have not been reported yet due to some issues in applying proxy-based losses in a hyperbolic space.
However, proxy-based losses are attractive for large-scale datasets since they have less training complexity.
To address these, this paper proposes the Combined Hyperbolic and Euclidean Soft Triple (CHEST) loss.
CHEST loss is composed of the proxy-based losses in hyperbolic and Euclidean spaces and the regularization loss based on hyperbolic hierarchical clustering.
We find that the combination of hyperbolic and Euclidean spaces improves DML accuracy and learning stability for both spaces.
Finally, we evaluate the CHEST loss on four benchmark datasets, achieving a new state-of-the-art performance.
\end{abstract}



\keywords{Deep metric learning \and Computer vision \and Image retrieval \and Feature extraction}

\section{Introduction}
The deep metric learning (DML) objective is to learn a mapping to an embedding space in which metrics such as distances between data represent semantic similarities.
Semantic similarities are based on class in most cases \cite{SoftTriple, MS, ProxyAnchor, MPAFam}.
The same class data are gathered nearby, while different class data are far away in the embedding space learned by DML.
This property allows neural networks to be applied to unseen data.
Hence, DML has been applied to various image tasks, such as few-shot learning \cite{FewShot}, face recognition \cite{Triplet, FaceRec}, anomaly detection \cite{RegAD}, and image retrieval \cite{CUB, InShop}.

The loss functions of DML can be categorized into two types of losses: pair-based and proxy-based losses.
Pair-based losses use the similarities between data \cite{Triplet, Npair, Contrastive, LiftedStructure, MS}.
In contrast, proxy-based losses use the similarity between data and the semantic center in embedding space \cite{ProxyNCA, ProxyAnchor, SoftTriple, MPAFam}.
According to these properties, pair-based losses require pair relationships, while proxy-based losses require semantic classes for each data.
The significant difference between pair-based and proxy-based losses is training complexity.
Training complexity has a significant impact on convergence speed and batch sampling \cite{SoftTriple,MPAFam}.
Proxy-based losses generally have less training complexity than pair-based losses \cite{ProxyAnchor, MPAFam}.
Hence, proxy-based loss can reduce the number of training steps and mitigate the effects of batch sampling \cite{SoftTriple}.

Conventional DML uses Euclidean space as an embedding space \cite{MS,ProxyAnchor}.
Afterward, a hyperbolic vision transformer (Hyp-ViT) \cite{HypViT} was proposed, which utilizes a hyperbolic space represented by the Poincar\'e ball model as its embedding space.
Hyperbolic space is attractive for DML because it can represent complex structures like tree structures \cite{HypEmb}.
The loss of Hyp-ViT is a pairwise cross-entropy loss, a type of pair-based loss \cite{HypViT}.
Then, HIER \cite{HIER} was also proposed, which is the regularization through unsupervised hierarchical clustering in hyperbolic space.
However, a proxy-based loss that is valid in hyperbolic space has not yet been proposed.
Proxy-based losses in hyperbolic spaces are more difficult to learn stable proxy learning than in Euclidean space.

The difficulty of applying proxy-based loss in hyperbolic space is caused by several reasons, such as the difference in similarity scales in hyperbolic and Euclidean spaces.
To address these problems, we propose the Combined Hyperbolic and Euclidean Soft Triple (CHEST) loss, which contributes to the stability of training in hyperbolic space DML.
CHEST loss is composed of SoftTriple loss \cite{SoftTriple} in both hyperbolic and Euclidean spaces.
CHEST loss regularizes proxies with the hierarchical clustering-based regularization. 
Furthermore, in order to stabilize the proxy learning, proxies are defined in Euclidean space and mapped to hyperbolic space using exponential mapping.
The main findings of our paper are the following:
\begin{itemize}
    \item The combination of losses in hyperbolic and Euclidean spaces improves DML accuracy and learning stability in both spaces by improving generalization bounds.
    \item CHEST loss outperforms state-of-the-art methods on four benchmark datasets.
\end{itemize}

\section{Related Work}
DML aims to learn an embedding space mapped by a neural network, where similar data are close to each other and dissimilar data are far away in the embedding space.
In this section, we introduce the loss functions for DML and hyperbolic DML.

\subsection{Deep Metric Learning Loss Functions}
DML loss function can be classified into two types of loss functions.
The first loss type is pair-based losses, and the second one is proxy-based losses.
Pair-based losses utilize the similarities between data.
Triplet loss \cite{Triplet} is calculated using pairs, including anchor, positive, and negative data.
In addition, other pair-based losses use all pairs in a batch \cite{LiftedStructure, Npair, MS,HypViT}. 
On the other hand, proxy-based losses utilize the similarities between data and proxies representing semantic centers for each class.
Proxy-based losses assume single or multiple proxies. 
ProxyNCA loss \cite{ProxyNCA} and ProxyAnchor loss \cite{ProxyAnchor} are based on a single proxy, and SoftTriple loss \cite{SoftTriple} and MPA loss \cite{MPAFam} are based on multiple proxies.

The important difference between pair-based and proxy-based loss is the complexity.
The space complexity of proxy-based losses is larger than that of pair-based losses.
In contrast, the learning complexity of proxy-based losses is proportional to dataset size, while that of pair-based losses grows in the cube of dataset size \cite{ProxyAnchor, MPAFam}.
The learning complexity affects the convergence speed for learning.
In particular, the learning complexity significantly affects when DML is applied to large-scale datasets.

\subsection{Hyperbolic Embeddings}
Hyperbolic space can embed the tree structures, and this space is attractive as a space to embed feature vectors \cite{HypEmb}.
According to this property, hyperbolic embedding has been proposed for various tasks \cite{HypEmb, HypNN, HypViT}.
Hyperbolic embedding was first proposed for NLP tasks \cite{HypEmb}.
Then, hyperbolic embedding was applied with image embedding tasks, such as few-shot and zero-shot tasks \cite{HypKernel, HypFew, HypImage}.

Hyperbolic embeddings have been applied to DML tasks \cite{HypViT}.
The network architectures for hyperbolic DML have a backbone network and mapping to hyperbolic space as a final layer \cite{HypViT}, the same as previous hyperbolic image embeddings \cite{HypImage}.
Hyp-ViT (Hyperbolic Vision Transformer) \cite{HypViT} uses pair-wise cross-entropy loss.
Additionally, proxy-based losses were reported to have less performance than pair-wise cross-entropy loss in hyperbolic space \cite{HypViT}.
After that, HIER (HIErarchical Regularization) \cite{HIER} was proposed as an unsupervised regularization loss for metric learning in hyperbolic space.

\section{Hyperbolic Proxy Deep Metric Learning}
This section proposes a Combined Hyperbolic and Euclidean Soft Triple (CHEST) loss.
Firstly, we introduce hyperbolic embedding, the same as in conventional studies \cite{HypEmb,HypViT}.
Then, we propose the network architecture and proxies for CHEST loss.
We also propose a similarity loss of CHEST loss at hyperbolic and Euclidean spaces.
Nextly, we introduce a hierarchical clustering-based regularization in a hyperbolic space to the CHEST loss.
Finally, we consider the complexity of the CHEST loss.

\subsection{Preliminary: Hyperbolic Embedding with Poincar\'{e} Ball}
The $n$-dimensional hyperbolic space $\mathbb{H}^{n}$ is a type of Riemannian manifold, and it has several isometric models.
Previous work has utilized the Poincar\'e ball model as a hyperbolic embedding in neural networks \cite{HypEmb, HypViT, HIER}.
In this paper, we also employ the Poincar\'e ball model.
Let $( \mathbb{D}^{n}_{\hypc}, g^{\mathbb{D}} )$ mean the $n$-dimensional Poincar\'e ball model, where $\hypc$ means a curvature hyperparameter.
The manifold $\mathbb{D}^{n}_{\hypc} = \{ \mathbf{x} \in \mathbb{R}^{n} | \hypc \| \mathbf{x} \| < 1 \}$ and Riemannian metric $g^{\mathbb{D}} = ( \lambda_{\mathbf{x}}^{\hypc} )^{2} g^{E} = ( \frac{2}{1-\hypc \left\| \mathbf{x} \right\|^{2}} )^{2} g^{E}$, where $g^{E}$ is the Euclidean metric.
$\lambda_{\mathbf{x}}^{\hypc}$ is the conformal factor, and this factor approaches infinity around the boundary of the Poincar\'e ball.
Hence, this property makes the space of Poincar\'e ball explosively large.

The data $\mathbf{u}, \mathbf{v} \in \mathbb{D}^{n}_{\hypc}$ in hyperbolic space can not be calculated using vector space algebraic operations.
The calculation of hyperbolic space needs to introduce the gyrovector space.
The addition operation in gyrovector space, M\"{o}bius addition, is defined as:
\begin{equation}
    \mathbf{u} \oplus_{\hypc} \mathbf{v} = \frac{\left(1 + 2 \hypc \left< \mathbf{u}, \mathbf{v} \right> + \hypc \left\| \mathbf{v} \right\|^{2} \right) \mathbf{u} + \left( 1 - \hypc \left\| \mathbf{u} \right\|^{2} \right) \mathbf{v}}{1 + 2 \hypc \left< \mathbf{u}, \mathbf{v} \right> + \hypc^{2} \left\| \mathbf{u} \right\|^{2} \left\| \mathbf{v} \right\|^{2}}.
\end{equation}
The distance metric in the Poincar\'e ball model using M\"{o}bius addition is defined as:
\begin{equation}
\label{HypDistance}
D_{H} \left( \mathbf{u}, \mathbf{v} \right) = \frac{2}{\sqrt{\hypc}} \arctanh \left( \sqrt{\hypc} \left\| - \mathbf{u} \oplus_{\hypc} \mathbf{v} \right\| \right).
\end{equation}
If $\hypc \rightarrow 0$, Equation~\eqref{HypDistance} is equivalent to Euclidean distance.

The outputs of general neural networks are in Euclidean space.
Euclidean space and the Poincar\'e ball model of hyperbolic space can be mapped to each other using mapping.
The mapping of Euclidean space to the Poincar\'e ball model is called exponential mapping.
On the other hand, the inverse mapping of the Poincar\'e ball model to Euclidean space is called the logarithm mapping.
In this paper, we use only exponential mapping.
The exponential mapping $\exp_{\mathbf{z}}: \mathbb{R}^{n} \rightarrow \mathbb{D}^{n}_{\hypc}$ for anchor $\mathbf{z} \in \mathbb{D}^{n}_{\hypc}$ is defined as
\begin{equation}
\exp_{\mathbf{z}} \left( \mathbf{x} \right) = \mathbf{z} \oplus_{\hypc} \left( \tanh \left( \sqrt{\hypc} \frac{\lambda_{\mathbf{z}}^{\hypc} \left\| \mathbf{x} \right\|}{2} \frac{\mathbf{x}}{\sqrt{\hypc} \left\| \mathbf{x} \right\|} \right) \right).
\end{equation}
Note that the exponential mapping when $\mathbf{z} = \mathbf{0}$ is defined as
\begin{equation}
\exp_{\mathbf{0}} \left( \mathbf{x} \right) = \tanh \left( \sqrt{\hypc} \left\| \mathbf{x} \right\| \right) \frac{\mathbf{x}}{\sqrt{\hypc} \left\| \mathbf{x} \right\|}.
\end{equation}

\subsection{Network Architecture and Proxies}
\begin{figure*}[tb]
    \begin{center}
        \includegraphics[width=\hsize]{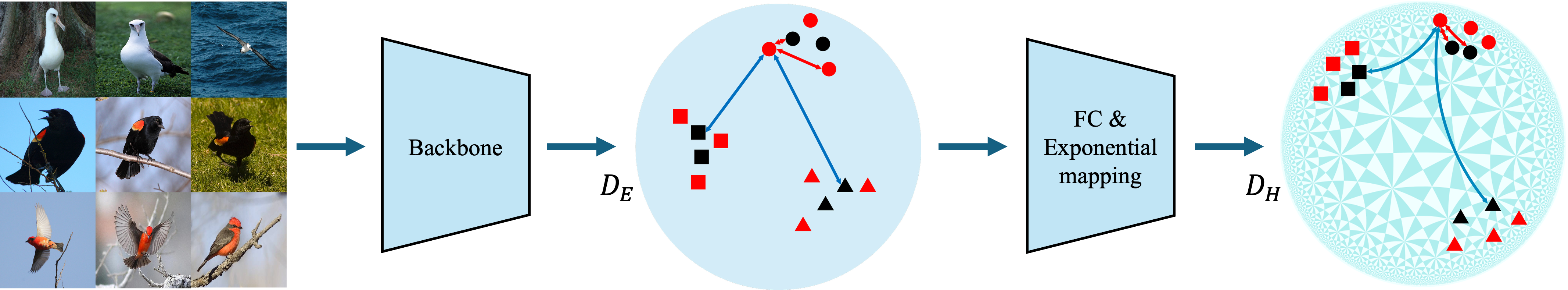}
    \end{center}
    \caption{
        Network architecture for the CHEST loss.
        $D_{E}$ means the dimension of Euclidean space, and $D_{H}$ means the dimension of hyperbolic space.
        Each point shape means a class.
        Red symbols are feature vectors and black symbols are proxies.
        The CHEST loss encourages red lines to be close and blue lines to be far away.
    }
    \label{fig:network_architecture}
\end{figure*}

We use an architecture just like a hyperbolic vision transformer (Hyp-ViT) that combines the vision transformer (ViT) with an exponential mapping network that maps Euclidean space to hyperbolic space \cite{HypViT}.
Figure~\ref{fig:network_architecture} shows the structure of the proposed network architecture and proxies.
Euclidean space dimension is $D_{E}$ and hyperbolic space dimension is $D_{H}$.
In addition, we introduce proxies for the CHEST loss to Euclidean space and map proxies to hyperbolic space.
This mapping is through a fully connected layer and exponential mapping, the same as the mapping of data from Euclidean to hyperbolic spaces.
They are optimized from the losses of both Euclidean and hyperbolic spaces in the CHEST loss, which will be discussed below.
This structure leads to stable learning of proxies.
These proxies have $K \geq 1$ proxies for each class.
It is also possible to define proxies separately in both spaces, but this may compromise the consistency of the proxies.
We denote $P_{E} = \{ \pe{i} \}_{i=1}^{C}$ in Euclidean space proxies and $P_{H} = \{ \ph{i} \}_{i=1}^{C}$ in hyperbolic space proxies, in which $C$ is the number of class, $\pe{i} = \{ \mpe{i}{j} \}_{j=1}^{K} \in \mathbb{R}^{K \times D_{E}}$, and $\ph{i} = \{ \mph{i}{j} \}_{j=1}^{K} \in \mathbb{R}^{K \times D_{H}}$.
Note that proxies in Euclidean space are not normalized.


\subsection{Combined Hyperbolic and Euclidean Soft Triple Loss}
Combined Hyperbolic and Euclidean Soft Triple (CHEST) loss aims to stabilize learning by utilizing a proxy-based loss in both hyperbolic and Euclidean spaces.
Hyperbolic space is attractive for DML because it can represent richer natural images.
However, distances in hyperbolic space are on a much larger scale than distances in Euclidean space, such as cosine similarity.
Additionally, gradients of distances in hyperbolic space increase rapidly near the boundary.
Most conventional proxy-based losses are based on cosine similarity, and it is known that large-scale similarities or distances negatively affect proxy-based loss \cite{SoftTriple, MPAFam}.
Additionally, the gradients of most conventional proxy-based losses are derived from the differences in similarity \cite{ProxyAnchor,MPAFam}.
Hence, large-scale similarities make DML challenging because it is difficult to get meaningful gradients due to the similarity differences in hyperbolic space between positive and negative data, which are much larger than small-scale similarities, such as cosine similarity.

CHEST loss designs to apply metric loss in both hyperbolic and Euclidean spaces to address this issue.
Models with CHEST loss learn embedding maps satisfying that similar data are near, and dissimilar data are far in both Euclidean and hyperbolic spaces. 
Even if the loss in hyperbolic space does not yield a beneficial gradient for learning, it is possible to obtain a beneficial gradient from the loss in Euclidean space.
Similarly, the inverse relation also holds.
Additionally, the CHEST loss is expected to be easier to learn, as it learns a mapping from learned vectors in Euclidean space to hyperbolic space, rather than exclusively using hyperbolic space loss.

Firstly, we introduce the notations used in this paper.
Let $X_E = \{ \xe{i} \}_{i=1}^{N}$ denote the feature vectors in Euclidean space, and $X_H = \{ \xh{i} \}_{i=1}^{N}$ denote the feature vectors in Hyperbolic space, where $N$ denotes the number of data.
$\mathcal{C} = \left\{ c_{i} \right\}_{i=1}^{N}$ denotes the corresponding class of data $\xe{i}$ and $\xh{i}$.
Then, CHEST loss uses similarities $\sime{\xe{i}}{\pe{c}}$ and $\simh{\xh{i}}{\ph{c}}$ between data and proxies in Euclidean and hyperbolic spaces.
In this paper, we use Euclidean distance $\dise{\cdot}{\cdot}$ and hyperbolic distance $\dish{\cdot}{\cdot}$, and similarities are defined as
\begin{align}
    \label{eq:sim}
    \sime{\xe{i}}{\pe{c}} &= - \sum_{k} \frac{\exp\left( - \frac{1}{\gamma_{E}} d_{E} \left( k \right) \right)}{\sum_{l} \exp \left( - \frac{1}{\gamma_{E}} d_{E} \left( l \right) \right)} d_{E} \left( k \right), \\
    \simh{\xh{i}}{\ph{c}} &= - \sum_{k} \frac{\exp\left( - \frac{1}{\gamma_{H}} d_{H} \left( k \right) \right)}{\sum_{l} \exp \left( - \frac{1}{\gamma_{H}} d_{H} \left( l \right) \right)} d_{H} \left( k \right),
\end{align}
where $d_{*} ( m ) = D_{*} (\mathbf{x}_{i}^{*}, \mathbf{p}_{c,m}^{*})$ and $*$ means $H$ or $E$.
$\gamma_{*}$ is a hyperparameter.
In addition, CHEST loss also uses the similarity $\simh{\sh{c}}{\ph{c}}$ between super-proxies and proxies in hyperbolic space.

CHEST similarity loss consists of two losses based on similarities in different spaces.
The first loss $\mathcal{L}_{H}$ is related to the similarities between data and proxies in hyperbolic space.
The second loss $\mathcal{L}_{E}$ is about the similarities between data and proxies in Euclidean space. 
Two losses have the same structure as the SoftTriple loss structure \cite{SoftTriple}.
For each data, CHEST similarity loss $\mathcal{L}_{sim} \left( \mathbf{x}_{i} \right)$ is defined as follows:
\begin{align}
    \label{eq:simloss}
    \mathcal{L}_{sim} \left( \mathbf{x}_{i} \right) &= \eta_{H} \mathcal{L}_{H} \left( \mathbf{x}_{i} \right) + \eta_{E} \mathcal{L}_{E} \left( \mathbf{x}_{i} \right) \nonumber \\
    &= - \eta_{H} \log \frac{\posf{H}{\xh{i}}{\ph{c_{i}}}}{\posf{H}{\xh{i}}{\ph{c_{i}}} + \sum_{c^{\prime} \neq c_{i}} \negf{H}{\xh{i}}{\ph{c^{\prime}}}} \nonumber \\
    & \quad - \eta_{E} \log \frac{\posf{E}{\xe{i}}{\pe{c_{i}}}}{\posf{E}{\xe{i}}{\pe{c_{i}}} + \sum_{c^{\prime} \neq c_{i}} \negf{E}{\xe{i}}{\pe{c^{\prime}}}} \nonumber \\
\end{align}
where $\posf{*}{\mathbf{x}^{*}}{p^{*}} = \exp \left( \lambda_{*} \left( S_{*} \left( \mathbf{x}^{*}, p^{*} \right) - \delta_{*} \right) \right)$ and $\negf{*}{\mathbf{x}^{*}}{p^{*}} = \exp \left( \lambda_{*} S_{*} \left( \mathbf{x}^{*}, p^{*} \right) \right)$, where $*$ means $H$ or $E$.
Let $\lambda_{*}$ denote the hyperparameters.
$\delta_{H}$ and $\delta_{E}$ denote the margin parameters in hyperbolic and Euclidean spaces, respectively.
Note that these losses, $\mathcal{L}_{H}$ and $\mathcal{L}_{E}$, can be replaced by the other losses, such as pair-based and other proxy-based losses.

From another point of view, CHEST similarity loss can be regarded as multi-task learning that trains the backbone from losses in different geometric spaces.
Multi-task learning is known to contribute to improved generalization performance \cite{MTL1, MTL2}.
Let the hypothesis sets defined by thresholds of empirical risk be denoted by $\mathcal{H}_{H}= \{ (\theta, \theta_{H}): \hat{R}_{H} (\theta, \theta_{H}) \leq \epsilon_{H} \}$ and $\mathcal{H}_{E}= \{ (\theta, \theta_{H}): \hat{R}_{E} (\theta) \leq \epsilon_{E} \}$, where $\theta$ and $\theta_{H}$ are the parameters of backbone and exponential mapping, respectively, $\hat{R}_{H}$ and $\hat{R}_{E}$ are empirical risks of hyperbolic and Euclidean spaces, and $\epsilon_{H}$ and $\epsilon_{E}$ are thresholds of hyperbolic and Euclidean spaces.
The hypothesis set of CHEST similarity loss is $\mathcal{H}_{CHEST} = \{ (\theta, \theta_{H}): \hat{R}_{H} \leq \epsilon_{H} \land \hat{R}_{E} \leq \epsilon_{E} \} = \mathcal{H}_{H} \cap \mathcal{H}_{E}$.
Therefore, the Rademacher complexity of CHEST similarity loss satisfies $\mathcal{R} (\mathcal{H}_{CHEST}) \leq \min ( \mathcal{R} (\mathcal{H}_{H}), \mathcal{R} (\mathcal{H}_{E}))$, where $\mathcal{R} (\cdot)$ is the Rademacher complexity with the same samples \cite{Rademacher2, Rademacher1}.
This relation leads to improving the generalization bounds.

\subsection{Hierarchical Clustering Regularization}
Proxy-based losses are computed based on similarities between data points and proxies \cite{SoftTriple, MPAFam}.
In general, proxies are learnable parameters for each class, and their learning is important for proxy-based losses.
Conventional multi-proxy losses perform regularization by encouraging proxies of the same class to move closer together.
CHEST loss utilizes hyperbolic and Euclidean spaces.
Hyperbolic space can embed tree structures \cite{HypEmb, HypViT}.
This property is attractive for deep metric learning.
To leverage this property, we incorporate hierarchical clustering-based regularization into the CHEST loss.

The regularization for the CHEST loss utilizes Hyperbolic Hierarchical Clustering (HypHC) with triplets sampling from proxies $P_{H}$ \cite{HypHC}.
HypHC is a kind of similarity-based hierarchical clustering.
However, since we cannot define static similarities between proxies, CHEST loss performs regularization based on the similarity between proxies based on the distance between them.
These similarities change at every step during training, but the CHEST similarity loss indirectly brings proxies of the same class closer together and separates proxies of different classes that are far away.
Proxies' similarities $\mathcal{S}_{i,j}$ between proxies $\mathbf{p}_{i}$ and $\mathbf{p}_{j}$ are defined as follows:
\begin{equation}
\mathcal{S}_{i,j} = \exp \left( - \dish{\mathbf{p}_{i}}{\mathbf{p}_{j}} \right).
\end{equation}
Additionally, HYPHC regularization samples triplets consisting of an anchor proxy $\mathbf{p}_{c,i}^{H}$, a same-class proxy as the anchor proxy $\mathbf{p}_{c,j}^{H}$, and a different-class proxy $\mathbf{p}_{c^{\prime},k}^{H}$, where $c \neq c^{\prime}$ and $i \neq j$.
For simplicity, we represent a triplet as $T = \{ t_{i} \}_{i=1}^{M} = \{ \mathbf{p}_{i,1}, \mathbf{p}_{i,2}, \mathbf{p}_{i,3} \}_{i=1}^{M}$, where $M$ is the number of triplets.
We also represent a distance as $d_{j,k} = \dish{\mathbf{p}_{i,j}}{\mathbf{p}_{i,k}}$.
HypHC regularization $\mathcal{L}_{HypHC}$ for each triplet is defined as follows \cite{HypHC}:
\begin{equation}
\label{eq:hyphc}
\mathcal{L}_{HypHC} \left( t_{i} \right) = \sum_{j < k} \mathcal{S}_{j,k} - \sum_{j<k} \mathcal{S}_{j,k} \frac{\exp \left( d_{j,k} / \gamma_{hyp} \right)}{\sum_{l<m} \exp \left( d_{l,m} / \gamma_{hyp} \right)},
\end{equation}
where $\gamma_{hyp}$ denotes the hyperparameter for the softmax function.
HypHC regularization encourages similar proxies to be on the same branch and dissimilar proxies to be on different branches.
As a result, same-class proxies are placed on the same branch, and different-class proxies are placed on different branches.
Finally, CHEST loss $\mathcal{L}$ is combined eq~\eqref{eq:simloss} and \eqref{eq:hyphc} as follows:
\begin{equation}
\label{eq:loss}
\mathcal{L} = \frac{1}{N} \sum_{i=1}^{N} \left( \eta_{H} \mathcal{L}_{H} \left( \mathbf{x}_{i} \right) + \eta_{E} \mathcal{L}_{E} \left( \mathbf{x}_{i} \right) \right) + \frac{\tau}{M} \sum_{i=1}^{M} \mathcal{L}_{HypHC} \left( t_{i} \right),
\end{equation}
where $\tau$ denotes the hyperparameter.

\subsection{Complexity Analysis}
The structure of CHEST similarity loss $\mathcal{L}_{sim}$ is basically the same as that of SoftTriple loss.
Hence, the training complexity of CHEST similarity loss is $O \left( N C K^{2}\right)$ \cite{MPAFam}.
Generally, the relationship of size is $N > C > K$.
Therefore, CHEST similarity loss has less training complexity than pair-based losses \cite{ProxyAnchor}.
On the other hand, HypHC regularization $\mathcal{L}_{HypHC}$ is based on triplet sampling from proxies in hyperbolic space.
The number of combinations of this sampling is $NK(NK-1)(NK-2)$.
The training complexity of HypHC regularization is $O ( C^{3} K^{3} )$.
Hence, a suitable triplet sampling strategy is essential to HypHC regularization.
The proposed sampling strategy is simple, but it can alleviate the complexity.
Hard negative sampling probably leads to efficient sampling.
However, it is difficult to apply to datasets with a huge number of classes due to spatial complexity.

The space complexity of the CHEST loss in the training process depends on the size of the distance matrix.
Let $B$ denote the batch size.
The space complexity of CHEST similarity loss $\mathcal{L}_{sim}$ is $O \left( B C K \right)$, as this loss calculates the distance between training data and proxies.
Additionally, the complexity of HypHC regularization $\mathcal{L}_{HypHC}$ is $O \left( 3M \right)$ since this loss only calculates the distance between triplets.
This complexity is much less than that of regularization for conventional multi-proxy losses, $O \left( C^{2} K^{2} \right)$.
Therefore, CHEST loss can apply with low space complexity.

\section{Experiments}
We demonstrate two experiments.
Firstly, we compare the CHEST loss with other DML methods.
Then, we provide an ablation study about the effect of loss components and the number of proxies.

\subsection{Datasets and Metrics}
We evaluate CHEST loss on CUB-200-2011 (CUB200) \cite{CUB}, Cars196 \cite{Cars}, In-shop Clothes Retrieval (In-shop) \cite{InShop}, and Stanford Online Products (SOP) \cite{LiftedStructure} datasets.
CUB200 \cite{CUB} contains 11,788 bird images in 200 classes.
Cars196 \cite{Cars} contains 16,185 car images of 196 classes.
In-shop \cite{InShop} contains 72,712 images in 7,986 categories.
SOP \cite{LiftedStructure} contains 120,053 images in 22,634 categories.
CUB200, Cars196, In-shop, and SOP datasets have an average of 58.9, 82.6, 9.1, and 5.3 data per class, respectively.
We use data split settings in conventional DML studies.

We use the Recall@k and MAP@R metrics as evaluation metrics.
The Recall@k metric is traditionally used for DML \cite{RealityCheck}. 
However, a comparison of the Recall@k metric may be meaningless for larger values of k.
Therefore, we provide the results of MAP@R metric.
We provide the results of both hyperbolic and Euclidean spaces.

\subsection{Implementation Details}
As in recent studies and Hyp-ViT, we utilize the Vision Transformer (ViT) architecture, pretrained on ImageNet-21k, as the backbone network \cite{ViT, ImageNet21K}.
We use two different ViT scales, ViT-Small (ViT-S) and ViT-Base (ViT-B), for comparability.
In all ViT, input images are divided into $16 \times 16$ patches.
Output vectors of ViT-S and ViT-B are 384 and 1024 dimensions, respectively.
The proxies and super-proxies dimensions $D_{E}$ in Euclidean space are the same as the ViT output dimensions.
Then, the outputs of exponential mapping, including a fully connected layer, are 384 and 512 dimensions for ViT-S and ViT-B, respectively.
Additionally, we compared CHEST-ViT to SoftTriple \cite{SoftTriple}, MPA \cite{MPAFam}, Hyp-ViT \cite{HypViT}, HIER \cite{HIER}, VPTSP-G \cite{VPTSP}, and RS@K \cite{RS} with various architectures. 
We use hyperbolic space curvature parameter $\hypc = 0.5$ and clipping radius $r=2.3$.
Data augmentation for training uses a random horizontal flip, random cropping of images, and scaling to $224 \times 224$ with bicubic interpolation.
Data transformation for testing involves scaling the smaller edges of an image to 256 using bicubic interpolation and center cropping to $224 \times 224$.

The optimization method uses the AdamW optimizer \cite{AdamW} for all experiments.
Since the optimal margin value may vary depending on the dataset, we use $\delta_{H} = \{1, 5, 10, 20\}$ and $\delta_{E} = \{ 1, 5, 10, 20 \}$.
We search for the combination of $\delta_{H}$ and $\delta_{E}$ that yields the highest accuracy in ViT-S, and use the same parameters in ViT-B.
We set $\gamma_E=\gamma_H=5$, $\lambda_{E}=\lambda_{H}=20$, and $\eta_{H}=\eta_{E}=1$ for the components of the CHEST similarity loss, Equations~\eqref{eq:sim} and \eqref{eq:simloss}.
Additionally, we also set $\gamma_{hyp}=1$ and $\tau=0.5$ for the HypHC regularization, as shown in Equations~\eqref{eq:hyphc} and \eqref{eq:loss}.
Table~\ref{tab:training_parameters} shows the other training parameters.
The number of triplets for HypHC regularization $M$ is determined by the number of classes in each dataset.
All experiments are performed on a single NVIDIA RTX A6000 with 48GB of GPU memory.
\begin{table*}[tbp]
    \centering{
        \begin{tabular}{c|c|c|c|c}
            \hline
            Parameters & CUB200 & Cars196 & In-shop & SOP \\
            \hline 
            Batch size & $200$ & $198$ & $100$ & $75$ \\
            Training steps & $120$ & $1800$ & $30000$ & $50000$ \\
            Learning rate & $3.0 \times 10^{-5}$ & $1.0 \times 10^{-5}$ & $1.0 \times 10^{-5}$ & $1.0 \times 10^{-5}$ \\
            Proxy learning rate & $1.0 \times 10^{-2}$ & $1.0 \times 10^{-2}$ & $1.0 \times 10^{-1}$ & $1.0 \times 10^{-1}$ \\
            Proxy num $K$ & $10$ & $10$ & $2$ & $2$ \\
            Regularization triplets $M$ & $100$ & $98$ & $3997$ & $11318$ \\
            \hline
        \end{tabular}
    }
    \caption{Training parameters}
    \label{tab:training_parameters}
\end{table*}

\subsection{Comparison of State-of-the-art Methods}
Tables~\ref{tab:Results_CUB}, \ref{tab:Results_Cars}, \ref{tab:Results_Inshop}, and \ref{tab:Results_SOP} show the comparison results on four benchmark datasets.
The best combination of margins in hyperbolic and Euclidean spaces is $(\delta_{H}, \delta_{E}) = (20, 1)$, $(1, 5)$, $(10, 5)$, and $(1, 5)$ for CUB200, Cars196, In-shop, and SOP, respectively.
\begin{table*}[tbp]
    \centering{
        \begin{tabular}{c|c|cccc}
            \hline
            \multirow{2}{*}{Methods} & \multirow{2}{*}{Arch (dim)} & \multicolumn{4}{c}{CUB200} \\
            \cline{3-6}
            & & R@1 & R@2 & R@4 & MAP@R \\
            \hline
            SoftTriple \cite{SoftTriple} & I (512) & $65.4$ & $76.4$ & $84.5$ & - \\
            MPA \cite{MPAFam} & I (512) & $69.1$ & $89.1$ & $86.3$ & $26.6$ \\
            \hline
            Hyp-ViT \cite{HypViT} & ViT-S (384) & $85.6$ & $91.4$ & $\mathbf{94.8}$ & - \\
            HIER \cite{HIER} & ViT-S (384) & $85.7$ & $91.3$ & $94.4$ & - \\
            VPTSP-G \cite{VPTSP} & ViT-S (384) & $\mathbf{86.6}$ & $\mathbf{91.7}$ & $\mathbf{94.8}$ & $52.7$ \\
            \rowcolor{gray!20} CHEST-ViT & ViT-S-E (384) & $86.0$ & $91.1$ & $94.2$ & $50.8$ \\
            \rowcolor{gray!20} CHEST-ViT & ViT-S-H (384) & $\mathbf{86.6}$ & $\mathbf{91.7}$ & $94.6$ & $\mathbf{53.9}$ \\
            \hline
            VPTSP-G \cite{VPTSP} & ViT-B (512) & $88.5$ & $\mathbf{92.8}$ & $\mathbf{95.1}$ & - \\
            \rowcolor{gray!20} CHEST-ViT & ViT-B-E (1024) & $88.7$ & $92.7$ & $95.0$ & $57.4$ \\
            \rowcolor{gray!20} CHEST-ViT & ViT-B-H (512) & $\mathbf{88.8}$ & $92.7$ & $94.8$ & $\mathbf{58.1}$ \\
            \hline
        \end{tabular}
    }
    \caption{
        Comparison results on the CUB200 dataset. 
        The rows with a gray background are the proposed method.
        Bold letters indicate the best performance for each architecture.
        ViT-S-E and ViT-B-E denote the results of Euclidean space, which are the output of ViT.
        ViT-S-H and ViT-B-H denote the results of hyperbolic space, which are the output of exponential mapping.
    }
    \label{tab:Results_CUB}
\end{table*}
\begin{table*}[tbp]
    \centering{
        \begin{tabular}{c|c|cccc}
            \hline
            \multirow{2}{*}{Methods} & \multirow{2}{*}{Arch (dim)} & \multicolumn{4}{c}{Cars196} \\
            \cline{3-6}
            & & R@1 & R@2 & R@4 & MAP@R \\
            \hline
            SoftTriple \cite{SoftTriple} & I (512) & $84.5$ & $90.7$ & $94.5$ & - \\
            MPA \cite{MPAFam} & I (512) & $87.1$ & $92.4$ & $95.5$ & $28.6$  \\
            \hline
            Hyp-ViT \cite{HypViT} & ViT-S (384) & $86.5$ & $93.3$ & $96.1$ & - \\
            HIER \cite{HIER} & ViT-S (384) & $88.3$ & $93.2$ & $96.1$ & - \\
            VPTSP-G \cite{VPTSP} & ViT-S (384) & $87.7$ & $93.3$ & $96.1$ & $28.9$ \\
            \rowcolor{gray!20} CHEST-ViT & ViT-S-E (384) & $88.5$ & $93.6$ & $96.5$ & $30.4$ \\
            \rowcolor{gray!20} CHEST-ViT & ViT-S-H (384) & $\mathbf{89.1}$ & $\mathbf{94.0}$ & $\mathbf{96.8}$ & $\mathbf{31.7}$ \\
            \hline
            RS@K \cite{RS} & ViT-B (512) & $89.5$ & $94.2$ & $96.6$ & - \\
            VPTSP-G \cite{VPTSP} & ViT-B (512) & $91.2$ & $95.1$ & $97.3$ & - \\
            \rowcolor{gray!20} CHEST-ViT & ViT-B-E (1024) & $91.9$ & $95.3$ & $97.4$ & $35.3$ \\
            \rowcolor{gray!20} CHEST-ViT & ViT-B-H (512) & $\mathbf{93.1}$ & $\mathbf{96.4}$ & $\mathbf{97.8}$ & $\mathbf{39.4}$ \\
            \hline
        \end{tabular}
    }
    \caption{
        Comparison results on the Cars196 dataset. 
        The rows with a gray background are the proposed method.
        Bold letters indicate the best performance for each architecture.
        ViT-S-E and ViT-B-E denote the results of Euclidean space, which are the output of ViT.
        ViT-S-H and ViT-B-H denote the results of hyperbolic space, which are the output of exponential mapping.
    }
    \label{tab:Results_Cars}
\end{table*}
\begin{table*}[tbp]
    \centering{
        \begin{tabular}{c|c|cccc}
            \hline
            \multirow{2}{*}{Methods} & \multirow{2}{*}{Arch (dim)} & \multicolumn{4}{c}{In-shop} \\ 
            \cline{3-6}
            & & R@1 & R@10 & R@20 & MAP@R \\
            \hline
            Hyp-ViT \cite{HypViT} & ViT-S (384) & $92.5$ & $98.3$ & $98.8$ & - \\
            HIER \cite{HIER} & ViT-S (384) & $92.8$ & $98.4$ & $\mathbf{99.0}$ & - \\
            VPTSP-G \cite{VPTSP} & ViT-S (384) & $91.2$ & $97.6$ & $98.4$ & - \\
            \rowcolor{gray!20} CHEST-ViT & ViT-S-E (384) & $93.4$ & $\mathbf{98.6}$ & $\mathbf{99.0}$ & $\mathbf{60.0}$ \\
            \rowcolor{gray!20} CHEST-ViT & ViT-S-H (384) & $\mathbf{93.5}$ & $\mathbf{98.6}$ & $\mathbf{99.0}$ & $59.8$ \\
            \hline
            VPTSP-G \cite{VPTSP} & ViT-B (512) & $92.5$ & $98.2$ & $98.9$ & - \\
            \rowcolor{gray!20} CHEST-ViT & ViT-B-E (1024) & $\mathbf{94.5}$ & $\mathbf{99.0}$ & $\mathbf{99.3}$ & $\mathbf{63.6}$ \\
            \rowcolor{gray!20} CHEST-ViT & ViT-B-H (512) & $\mathbf{94.5}$ & $98.8$ & $99.2$ & $63.3$ \\
            \hline
        \end{tabular}
    }
    \caption{
        Comparison results on the In-shop dataset. 
        The rows with a gray background are the proposed method.
        Bold letters indicate the best performance for each architecture.
        ViT-S-E and ViT-B-E denote the results of Euclidean space, which are the output of ViT.
        ViT-S-H and ViT-B-H denote the results of hyperbolic space, which are the output of exponential mapping.
    }
    \label{tab:Results_Inshop}
\end{table*}
\begin{table*}[tbp]
    \centering{
        \begin{tabular}{c|c|cccc}
            \hline
            \multirow{2}{*}{Methods} & \multirow{2}{*}{Arch (dim)} & \multicolumn{4}{c}{SOP} \\ 
            \cline{3-6}
            & & R@1 & R@10 & R@100 & MAP@R \\
            \hline
            SoftTriple \cite{SoftTriple} & I (512) & $78.3$ & $90.3$ & $95.9$ & - \\
            MPA \cite{MPAFam} & I (512) & $78.1$ & $90.1$ & $95.6$ & $50.2$ \\
            \hline
            Hyp-ViT \cite{HypViT} & ViT-S (384) & $85.9$ & $94.9$ & $\mathbf{98.1}$ & - \\
            HIER \cite{HIER} & ViT-S (384) & $86.1$ & $\mathbf{95.0}$ & $98.0$ & - \\
            VPTSP-G \cite{VPTSP} & ViT-S (384) & $84.4$ & $93.6$ & $97.3$ & - \\
            \rowcolor{gray!20} CHEST-ViT & ViT-S-E (384) & $86.3$ & $94.7$ & $97.5$ & $63.7$ \\
            \rowcolor{gray!20} CHEST-ViT & ViT-S-H (384) & $\mathbf{86.5}$ & $94.7$ & $97.6$ & $\mathbf{64.2}$ \\
            \hline
            RS@K \cite{RS} & ViT-B (512) & $88.0$ & $\mathbf{96.1}$ & $\mathbf{98.6}$ & - \\
            VPTSP-G \cite{VPTSP} & ViT-B (512) & $86.8$ & $95.0$ & $98.0$ & - \\
            \rowcolor{gray!20} CHEST-ViT & ViT-B-E (1024) & $\mathbf{88.2}$ & $95.8$ & $98.1$ & $\mathbf{67.3}$ \\
            \rowcolor{gray!20} CHEST-ViT & ViT-B-H (512) & $88.0$ & $95.6$ & $97.9$ & $67.1$ \\
            \hline
        \end{tabular}
    }
    \caption{
        Comparison results on the SOP dataset. 
        The rows with a gray background are the proposed method.
        Bold letters indicate the best performance for each architecture.
        ViT-S-E and ViT-B-E denote the results of Euclidean space, which are the output of ViT.
        ViT-S-H and ViT-B-H denote the results of hyperbolic space, which are the output of exponential mapping.
    }
    \label{tab:Results_SOP}
\end{table*}
The results of the hyperbolic space of CHEST-ViT outperformed the state-of-the-art methods in R@1 for all datasets.
On the other hand, the results of Euclidean space outperformed those of hyperbolic space in some cases.
The results of both spaces 
These results suggest that both losses, $\mathcal{L}_{H}$ and $\mathcal{L}_{E}$, in hyperbolic and Euclidean spaces have a positive effect on the backbone.

Compared to Hyp-ViT, CHEST-ViT had higher R@1 for all datasets.
The large difference between Hyp-ViT and CHEST-ViT is the size of the generalization bound.
CHEST-ViT has a smaller generalization bound than Hyp-ViT because CHEST-ViT is optimized in both hyperbolic and Euclidean space.
Hence, CHEST-ViT is expected to converge more easily to weights with higher generalization performance.
Additionally, CHEST loss is suitable to learn on large-scale datasets because it has lower training complexity than Hyp-ViT loss \cite{HypViT,ProxyAnchor}.
Therefore, CHEST-ViT outperforms Hyp-ViT in terms of performance and has higher applicability across various datasets.

\begin{figure}[tbp]
    \begin{center}
        \includegraphics[width=\hsize]{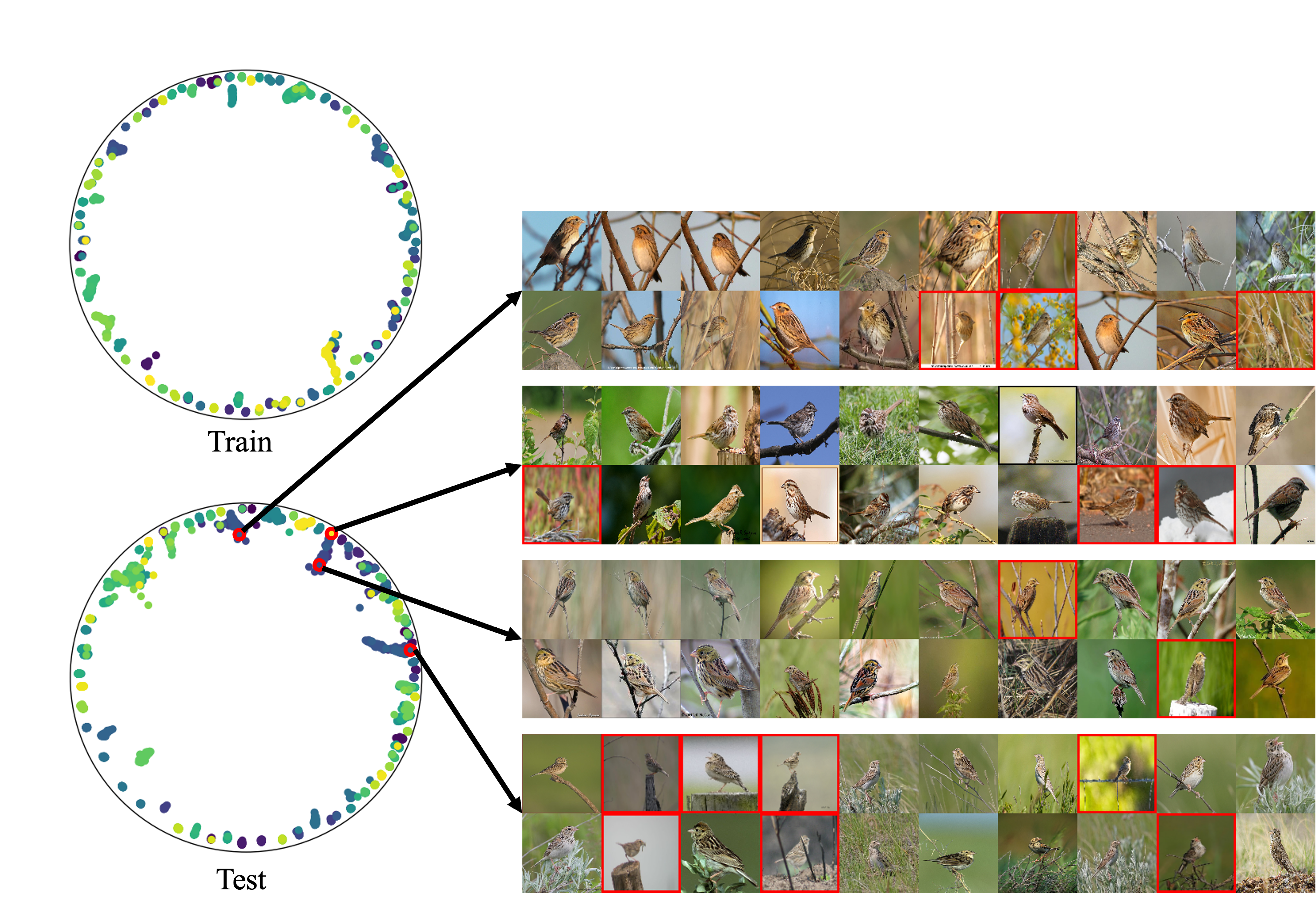}
    \end{center}
    \caption{
        Distribution of embeddings on CUB200 train and test datasets and examples of retrieval results on the CUB200 dataset.
        The red circles in the distribution represent the retrieval query of examples.
        Four examples of retrieval show.
        Each retrieval result has two rows, including one query and 19 retrieval results.
        The top left images are query images, and the other images are arranged in order of similarity from top left to top right and bottom left to bottom right. 
        A red frame means the retrieved image is a different class image.
    }
    \label{fig:CUB_map_retrieval}
\end{figure}

Figure~\ref{fig:CUB_map_retrieval} shows the embeddings of the CUB200 train and test datasets on the Poincar\'e disk and examples of the retrieval results.
In this figure, CHEST with ViT-B embedded data in a hyperbolic embedding space; after that, they were compressed in dimension with UMAP using a hyperbolic metric.
These retrieval examples can be retrieved semantic similarity data such as shape, color, and pattern, regardless of backgrounds.
Examples in close similarity in the embedding space share similar features, while examples far apart have different features.

\subsection{Ablation Study}
We also provide the ablation study for CHEST loss with ViT-S on Cars196 and In-shop datasets.
This ablation study validates the impacts of hyperbolic space loss ($\eta_{H}$), Euclidean space loss ($\eta_{E}$), multi-proxies ($K$), and HypHC regularization ($\tau$). 
The other settings are the same as the comparison of state-of-the-art methods.
The ablation settings are combinations of $\eta_{H}=0$, $\eta_{E}=0$, $\tau=0$, and $K=1$.
We set $\eta_{H}=\{0,1\}$, $\eta_{E}=\{0,1\}$, and $\tau=\{0,0.5\}$.
We also set $K = \{1, 10\}$ for Cars196 dataset and $K = \{ 1, 2 \}$ for In-shop dataset.
When $\eta_{H} = 1$ and $\eta_{E}=1$, the Euclidean loss is computed with the backbone output, $D_{E}$ dimension vectors.
On the other hand, when $\eta_{H}=0$ and $\eta_{E}=1$, Euclidean loss is computed with the output of CHEST-ViT without exponential mapping.

\begin{figure}[tbp]
    \begin{center}
        \begin{minipage}[tbp]{0.48\hsize}
            \begin{center}
                \includegraphics[width=0.9\hsize]{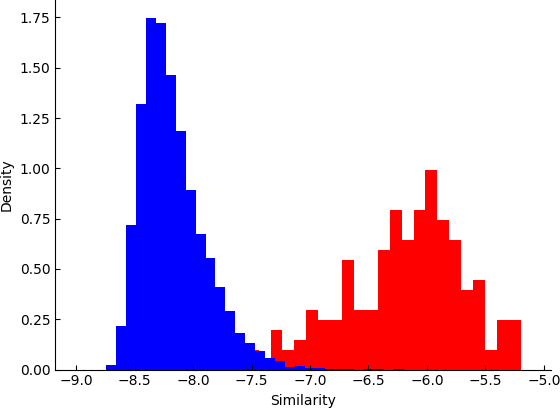}
            \end{center}
        \end{minipage}
        \begin{minipage}[tbp]{0.48\hsize}
            \begin{center}
                \includegraphics[width=0.9\hsize]{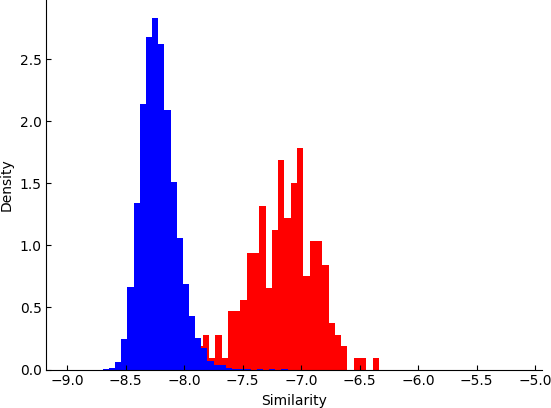}
            \end{center}
        \end{minipage}
    \end{center}
    \caption{
        The histogram of similarities between batch data and classes.
        The left histogram result is without Euclidean space loss, and the right histogram result is of the combined hyperbolic and Euclidean loss (Hyperbolic + Euclidean).
        The blue area represents the similarities to negative classes, and the red area represents the similarities to positive class.
    }
    \label{fig:SimHist}
\end{figure}

Table~\ref{tab:Ablation_Cars} and \ref{tab:Ablation_Inshop} show the ablation results on Cars196 and In-shop datasets, respectively.
\begin{table*}[tbp]
    \centering{
        \begin{tabular}{l|c|c|c|c|c||c}
            \hline
            \multicolumn{1}{c|}{Losses} & Space & $\eta_{H}$ & $\eta_{E}$ & $K$ & $\tau$ & R@1 \\ 
            \hline
            Hyperbolic & H & $1$ & $0$ & $1$ & $0.0$ & $84.5$ \\
            Euclidean & E & $0$ & $1$ & $1$ & $0.0$ & $83.1$ \\
            Hyperbolic + Euclidean & E & $1$ & $1$ & $1$ & $0.0$ & $88.3$ \\
            Hyperbolic + Euclidean & H & $1$ & $1$ & $1$ & $0.0$ & $88.8$ \\
            Hyperbolic + Multi & H & $1$ & $0$ & $10$ & $0.0$ & $85.2$ \\
            Euclidean + Multi & E & $0$ & $1$ & $10$ & $0.0$ & $83.4$ \\
            Hyperbolic + Euclidean + Multi & E & $1$ & $1$ & $10$ & $0.0$ & $88.4$ \\
            Hyperbolic + Euclidean + Multi & H & $1$ & $1$ & $10$ & $0.0$ & $89.0$ \\
            Hyperbolic + Multi + HypHC & H & $1$ & $0$ & $10$ & $0.5$ & $85.0$ \\
            CHEST loss & E & $1$ & $1$ & $10$ & $0.5$ & $88.5$ \\
            CHEST loss & H & $1$ & $1$ & $10$ & $0.5$ & $\mathbf{89.1}$ \\
            \hline
        \end{tabular}
    }
    \caption{
        Ablation results on the Cars196 dataset.
        Hyperbolic is the baseline method, and this method only calculates $\mathcal{L}_{H}$.
        Euclidean and Regularization add $\mathcal{L}_{E}$ and $\mathcal{L}_{HypHC}$ to the baseline method, respectively.
        Multi denotes it has multiple proxies per class.
        CHEST loss has all components.
        The column of space denotes the output space.
        H denotes the hyperbolic space, and E denotes the Euclidean space.
        The dimension of both spaces is 384.
    }
    \label{tab:Ablation_Cars}
\end{table*}
\begin{table*}[tbp]
    \centering{
        \begin{tabular}{l|c|c|c|c|c||c}
            \hline
            \multicolumn{1}{c|}{Losses} & Space & $\eta_{H}$ & $\eta_{E}$ & $K$ & $\tau$ & R@1 \\ 
            \hline
            Hyperbolic & H & $1$ & $0$ & $1$ & $0.0$ & $91.4$ \\
            Euclidean & E & $0$ & $1$ & $1$ & $0.0$ & $92.0$ \\
            Hyperbolic + Euclidean & E & $1$ & $1$ & $1$ & $0.0$ & $93.2$ \\
            Hyperbolic + Euclidean & H & $1$ & $1$ & $1$ & $0.0$ & $93.3$ \\
            Hyperbolic + Multi & H & $1$ & $0$ & $2$ & $0.0$ & $91.2$ \\
            Euclidean + Multi & E & $0$ & $1$ & $2$ & $0.0$ & $91.9$ \\
            Hyperbolic + Euclidean + Multi & E & $1$ & $1$ & $2$ & $0.0$ & $93.2$ \\
            Hyperbolic + Euclidean + Multi & H & $1$ & $1$ & $2$ & $0.0$ & $93.4$ \\
            Hyperbolic + Multi + HypHC & H & $1$ & $0$ & $2$ & $0.5$ & $91.1$ \\
            CHEST loss & E & $1$ & $1$ & $2$ & $0.5$ & $93.4$ \\
            CHEST loss & H & $1$ & $1$ & $2$ & $0.5$ & $\mathbf{93.5}$ \\
            \hline
        \end{tabular}
    }
    \caption{
        Ablation results on the In-shop dataset.
        Hyperbolic is the baseline method, and this method only calculates $\mathcal{L}_{H}$.
        Euclidean and Regularization add $\mathcal{L}_{E}$ and $\mathcal{L}_{HypHC}$ to the baseline method, respectively.
        Multi denotes it has multiple proxies per class.
        CHEST loss has all components.
        The column of space denotes the output space.
        H denotes the hyperbolic space, and E denotes the Euclidean space.
        The dimension of both spaces is 384.
    }
    \label{tab:Ablation_Inshop}
\end{table*}
The hyperbolic space of CHEST loss was the highest R@1 for all settings.
The Euclidean space loss had the most significant impact on performance.
On the other hand, the HypHC regularization and multi-proxies were less effective than Euclidean space loss.
However, each factor was better than without each factor.
In addition, the results without Euclidean space loss showed a decrease in R@1 in the latter half of the training.
Figure~\ref{fig:SimHist} shows the histogram of similarities between batch data and classes at 1800 training steps on Cars196.
The only hyperbolic space loss tends to be getting closer between data and the positive class than the combined hyperbolic and Euclidean loss.
As a result, when the loss is only in hyperbolic space, the loss in hyperbolic space is made smaller than the combined hyperbolic and Euclidean loss.
On the other hand, the combined hyperbolic and Euclidean loss was much higher R@1 than the loss in hyperbolic space only.
Therefore, the loss in only hyperbolic space would be overfitting.
The combined hyperbolic and Euclidean loss has a tighter generalization bound than the single loss in hyperbolic or Euclidean spaces.
Thus, the losses in hyperbolic and Euclidean spaces perform like regularizers for each other, preventing overfitting and improving the stability and accuracy of learning.


\section{Conclusion}
We have proposed the Combined Hyperbolic and Euclidean Soft Triple (CHEST) loss.
CHEST loss combines the losses in hyperbolic space and Euclidean space.
Proxies are regularized based on hierarchical clustering to utilize the property of hyperbolic space.
CHEST showed better performance than the other state-of-the-art methods.
We found that a combination of losses in hyperbolic and Euclidean spaces leads to higher performance and higher learning stability in both spaces.
In addition, CHEST loss has the properties of proxy-based losses and is expected to reduce training complexity.
Therefore, CHEST loss would require fewer training steps when working with large-scale datasets.
Finally, CHEST has many hyperparameters, and some datasets should have a more optimal set of parameters.

\bibliographystyle{ieeetr}
\bibliography{references}
\end{document}